\documentclass[letterpaper, 10 pt, conference]{ieeeconf}  

\IEEEoverridecommandlockouts                              

\usepackage{graphicx}
\usepackage{xcolor}
\usepackage{amsmath}

\title{\LARGE \bf
Visual-Tactile Multimodality for Following Deformable Linear Objects Using Reinforcement Learning
}

\author{Leszek Pecyna$^{1}$, Siyuan Dong$^{2}$ and Shan Luo$^{3}$ 
\thanks{This work was supported by the EPSRC project ``ViTac: Visual-Tactile Synergy for Handling Flexible Materials" (EP/T033517/1).}
\thanks{$^{1}$Leszek Pecyna is with the smARTLab, Department of Computer Science,
        University of Liverpool, Liverpool L69 3BX, United Kingdom. Email:
        {\tt\small l.pecyna@liverpool.ac.uk}}%
\thanks{$^{2}$Siyuan Dong is with the Paul G. Allen School of Computer Science \& Engineering, University of Washington, Seattle, WA, USA. Email:
        {\tt\small siyuandong.bme@gmail.com}}%
\thanks{$^{3}$Shan Luo is with the Department of Engineering, King’s College London,
        London WC2R 2LS, United Kingdom. Email:
        {\tt\small shan.luo@kcl.ac.uk}}%
}

\begin{document}

\maketitle
\thispagestyle{empty}
\pagestyle{empty}

\begin{abstract}

Manipulation of deformable objects is a challenging task for a robot. It will be problematic to use a single sensory input to track the behaviour of such objects: vision can be subjected to occlusions, whereas tactile inputs cannot capture the global information that is useful for the task. In this paper, we study the problem of using vision and tactile inputs together to complete the task of following deformable linear objects, for the first time. We create a Reinforcement Learning agent using different sensing modalities and investigate how its behaviour can be boosted using visual-tactile fusion, compared to using a single sensing modality. To this end, we developed a benchmark in simulation for manipulating the deformable linear objects using multimodal sensing inputs. The policy of the agent uses distilled information, e.g., the pose of the object in both visual and tactile perspectives, instead of the raw sensing signals, so that it can be directly transferred to real environments. In this way, we disentangle the perception system and the learned control policy. Our extensive experiments show that the use of both vision and tactile inputs, together with proprioception, allows the agent to complete the task in up to 92\% of cases, compared to 77\% when only one of the signals is given. Our results can provide valuable insights for the future design of tactile sensors and for deformable objects manipulation.

\end{abstract}

\section{Introduction}

Humans and animals explore and interact with their environment through a variety of senses of a different modality. In some cases, we are able to observe the integration of different modalities when one signal affects the perception of other sensory inputs.
An example of such multi-modal influence is the McGurk effect \cite{mcgurk1976hearing}, in which humans' perception of particular sounds is affected by visual cues. 
Touch and vision are especially used by humans during object identification and manipulation. This can be seen in neuropsychological studies on fMRI data, which shows that both visual and haptic signals are processed in a cross-modal fashion during some of these tasks \cite{james2002haptic,blake2004neural}.

In contrast, most of the artificial systems are based on a single modality when performing their tasks and often different types of algorithms are developed to approach particular modalities. As robots are operating in a more complex and dynamic environments, it can be expected that the usage of variety of sensing modalities will play more important role for them \cite{luo2021vitac}.

One of the situations when the environment becomes more complicated, and multimodal perception might help in better comprehension of it, is manipulation of flexible objects. 
As such, contour following of Deformable Linear Objects (DLOs) is a common task performed by humans, e.g., cable following. We perform this by grasping a cable between the thumb and the forefinger and slide the fingers to the target position \cite{she2021cable}. This task is quite common in our daily life, for example, when untangling the cables or when following the cable to find its plug-end.

Cable following can be a challenging task for artificial systems as the cable shape is changing dynamically while the gripper is sliding. Moreover, different cables/ropes can be characterized by different stiffness and friction, and their starting shape might be complex and undetermined (kinks, intersections, etc.). Due to these challenges, most of the research concerning DLO manipulation uses some additional constraints, e.g., a object is placed on a table \cite{ yan2020self, zhu2018dual}.

\begin{figure}
	\centering
	\includegraphics[width = 1\columnwidth]{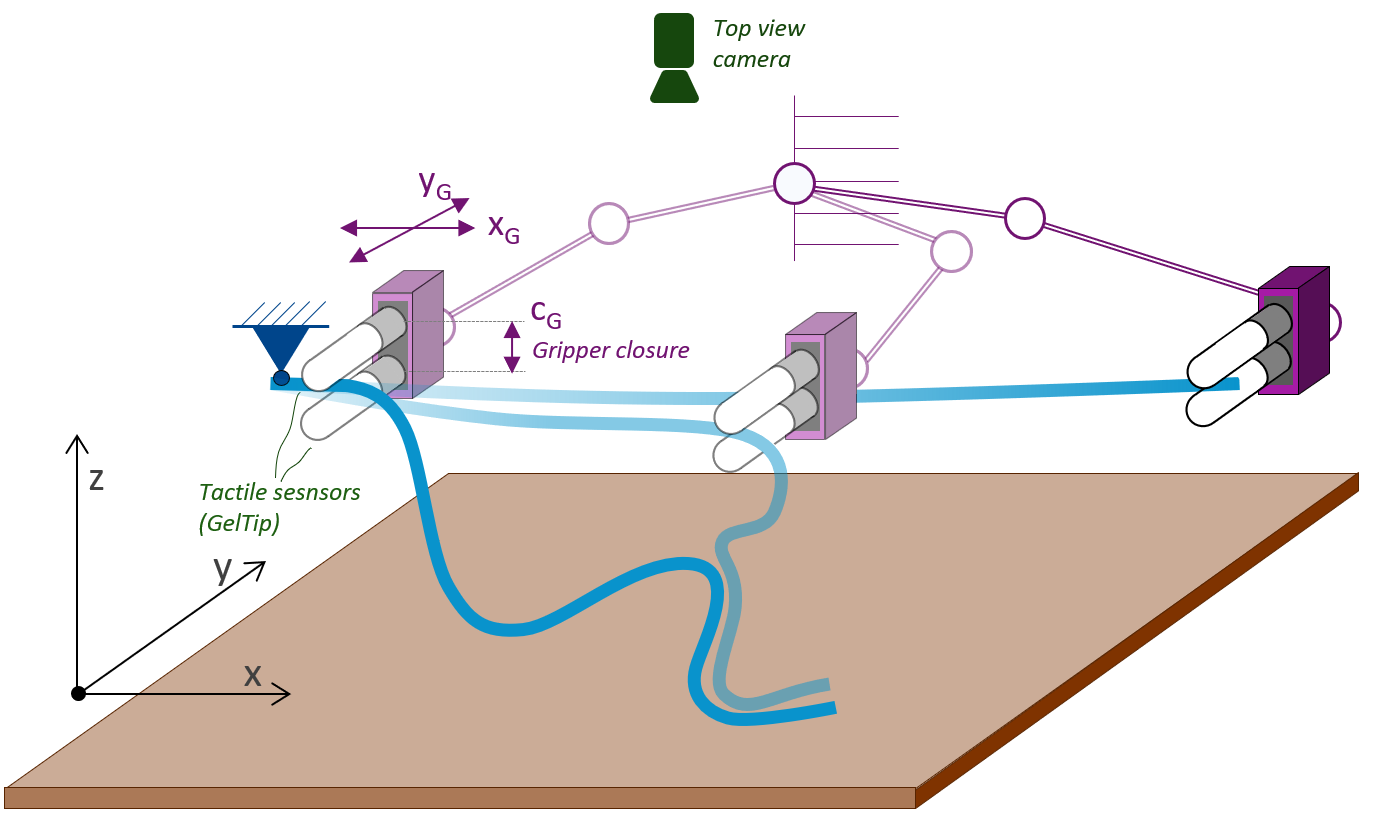}
	\caption{Manipulation goal: the gripper starts from the fixed end of the rope/cable (on the left) and follows it, up to its tail end (on the right).}
	\label{fig:Task}
\end{figure}

Training a Reinforcement Learning (RL) agent in a simulated environment, in many aspects, is a desirable approach as the environment can be explored through an extensive number of episodes without possibility of damaging a robot. However, simulation of sliding and realistic grasping is a challenging task itself. For this reason, most of prior works on flexible object manipulation utilize a firm grasp, i.e., the section or a point of the object is fixed to the gripper and cannot move in relation to it \cite{li2015regrasping, pmlr-v155-lin21a}. 
In our work, we cannot use such an approach as the gripper is sliding along the object, instead we aim to modulate the grasping force while the hand is moving.

In this paper, we create an RL agent for a cable/rope following task in a simulated environment and investigate how its behaviour can be boosted using visual-tactile fusion, compared to using a single sensing modality. As shown in Fig.~\ref{fig:Task}, we have both visual and tactile perspectives of the state of the deformable linear object in the gripper. The robot agent's goal is to pick up the object at its fixed end and follows it up to its tail end. 
We chose this task, as sliding along DLOs is not well explored, especially when it comes to simulation, and it is a good candidate for research concerning visual-tactile synergy as both of the signals can provide useful information to complete the task.


To the best of our knowledge, this is the first study to use both vision and tactile inputs for the task of cable following. We used distilled information as the observations of the agent, e.g., the object pose in both visual and tactile perspectives, instead of the raw visual images or tactile data. In this way, the trained agent can be directly transferred to real environments, and the learned control policy can be disentangled from the perception system. Through our extensive experiments, we find that when both vision and tactile inputs, together with proprioception, are used, the agent can complete the task (reach the end of the cable and hold it) in up to 92\% of cases, compared to the best result of 77\% with a single sensory input used (for vision); and when two signals were used -- 89\% (for vision and proprioception).


\section{Related Work} \label{related}

\subsection{Visual-Tactile Multimodality}

Vision and touch are two main important senses used for object manipulation. They have been widely used in robotics but, in most of the cases, with only one sensory input used \cite{luo2017robotic,luo2021vitac}. In the resent years, there have been several studies aiming to combine both of these inputs. Many of them concentrate on sensing, rather than on manipulation, like feature sharing or feature extraction \cite{yuan2017connecting, luo2018vitac}.
When it comes to object manipulation itself, even when both of the senses are utilised, in many cases, one input type supports another one, and is used in a specific sub-task, e.g., tactile sensing can help to verify the contact with the object (several examples of that are provided in \cite{luo2017robotic}). Hence, they are not used simultaneously together.

It seems to be less common to extract features from the sensors and use them together in the control scheme. In \cite{bekiroglu2013probabilistic} multiple sensory inputs are integrated in a grasping stability task (of mugs and bottles). In \cite{hebert2011fusion} in-hand object location is estimated from joint sensors, and \cite{bimbo2015global} covers object pose estimation. The work presented herein follows the idea where the extracted features are used simultaneously together.

\subsection{DLO Following}
There have been several studies concerning contour following of rigid objects that utilise vision \cite{lange1998predictive} or tactile sensing \cite{chen1995edge, ward2018tactip}.
From the point of view of flexible object following, that utilise tactile sensors, there are two works \cite{hellman2017functional} and \cite{she2021cable}. The first \cite{hellman2017functional} proposes a reinforcement learning approach to close a deformable ziplock bag using BioTac sensors. The robot grasps and follows the edge of the bag using a constant grasping force. The authors used Contextual Multi-Armed Bandits (C-MAB) RL algorithm to train the robot to close the bag in a discreet time steps with a maximum velocity of 0.5 cm/s (trapezoidal velocity profiles). 
The second \cite{she2021cable}, which is the most relevant to our work from the point of view of the task, presents a control framework that uses a real-time tactile feedback from a GelSight sensor \cite{yuan2017gelsight} to accomplish the task of cable following. To achieve that, the authors designed a parallel gripper with a servo motor actuator. Also, in their study, only the tactile signal was used, but the gripping force was modulated. The authors did not use the RL approach in that case, instead they used two controllers: PD -- for cable grip control, and LQR -- for cable pose control. Compared to the decoupled controllers in~\cite{she2021cable}, in this work we control both the gripping force and the end-effector pose simultaneously using the RL policy. 

\section{Problem Statement}

The goal of the presented DLO following task is to grasp the cable/rope at the beginning end with the gripper, and follow it -- using an appropriate grasping force -- to its tail end. The task should finish by holding the object close to its finishing end. The beginning of the rope is firmly held by the second gripper (it is attached to a point in space in the simulator). The problem is illustrated in Fig. \ref{fig:Task}.

In many aspects, this task is similar to the one presented in \cite{she2021cable}. There are, however, many differences: we performed the training in a simulator (we are planning to test our model on a real platform in the next stages of our work); we use data from both vision and tactile sensors (in \cite{she2021cable} only tactile signal was used); our model uses an RL algorithm (compared to decoupled PD and LQR controllers in \cite{she2021cable}). Also, we do not perform re-grasping procedure\footnote{This procedure was not not part of the controller task in \cite{she2021cable} and it was used when the gripper recognises it is loosing the cable and when the robot reaches its workspace limits.}, instead, we finish training episode when the DLO falls from the gripper. We assume no plug at the end of the cable (which allowed cable-end recognition by the tactile sensor in \cite{she2021cable}). Hence, in our case, we expect the vision to play a principal role in the object-end identification. 
As our task is conducted in the simulator, the parameters we chose can make the object properties correspond to those of a cable or a rope, which can be much softer than the cable used in \cite{she2021cable}.

Our model of DLO manipulation is defined as finite-horizon, discounted Markov decision process (MDP) represented by a tuple of ($\mathcal{S}$, $\mathcal{A}$, $p$, $\mathcal{R}$). 
The state space $\mathcal{S}$ and action space $\mathcal{A}$ are assumed to be continuous. State transition probability $p$, represents the probability density of the next state $s_{t+a} \in \mathcal{S}$ given the current state $s_t \in \mathcal{S}$ and action $a_t \in \mathcal{A}$. $\mathcal{R}$ is an immediate reward emitted by the environment on each transition. 
The details about the agent's actions and state space (observations), as well as, the definition of a reward function implementation are provided in the next section.

\section{Methodology}

In this section we present our methodology for implementation and usage of an RL agent to solve the task of DLO following. We first describe the model, its architecture, what observations it uses and what actions the agent can make. Next, we present the reward function that is set to promote the behaviour of reaching the object-end and staying there. Finally, we explain how our model performance is evaluated.

\subsection{Agent Description}

In our study we use Soft Actor-Critic (SAC) \cite{haarnoja2018soft, haarnoja2018soft2} that provides state-of-the-art performance in continuous control tasks (like robotic manipulation). 
SAC combines sample efficient off-policy method with ability to operate in a continuous action and state spaces. 


\subsubsection{Model Architecture}

\begin{figure}
	\centering
	\includegraphics[trim=0 0 0 0, clip, width = 0.64\columnwidth]{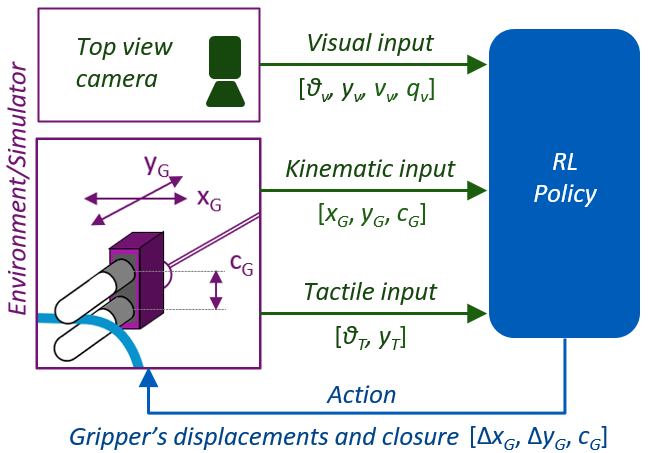}
	\caption{DLO following task with the RL policy.}
	\label{fig:policy}
\end{figure}

The model is composed of an actor network and a critic which is made of two Q-value networks (to combat the problem of overestimation of Q-values).

Both Q-value networks and the policy network are MLPs with two hidden layers of 1,024 neurons with ReLU activation function.
The actor takes as an input the state and outputs the mean and covariance for the Gaussian distribution that represents the policy \cite{haarnoja2018soft2}. From that the action is sampled.
The Q-value network input is made of actions together with observation space and produces single value (Q-value). The model's general scheme of interaction with the environment can be seen in Fig.~\ref{fig:policy}.

\subsubsection{Observations}
\begin{figure}
	\centering
	\includegraphics[trim=0 10 0 10, clip,width = 0.83\columnwidth]{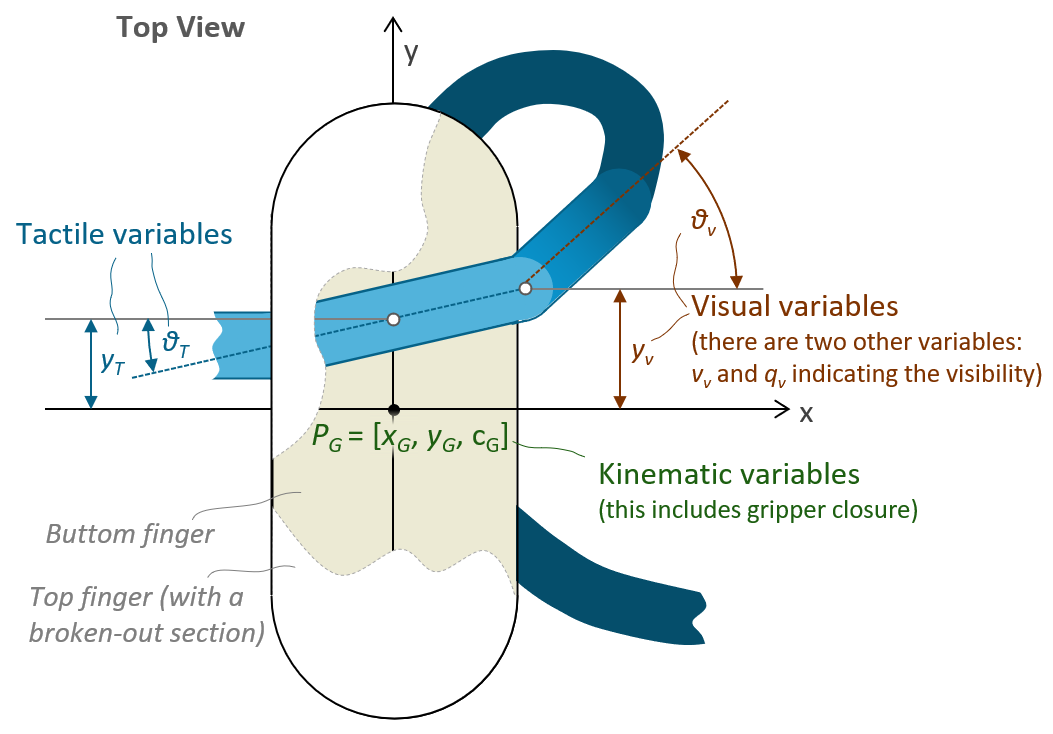}
	\caption{Illustration of state variables (observations) available for the model.}
	\label{fig:State1}
\end{figure}

In general, observations we use in our model can be divided in three categories as shown in Fig.~\ref{fig:State1}.
\begin{itemize}
    \item Kinematic (proprioceptive): it provides the information about position of the gripper in the space ($x$ and $y$ coordinates) and its closure state (variable from 0 to 1, where 0 corresponds to the situation where gripper is fully open and 1 where it is fully closed). It is an array of 3 components:
    $O_G = [x_G, y_G, c_G]$.
    \item Tactile: Described more in Section \ref{tactile}, it is composed of an angle and position of the DLO in relation to the gripper: $O_T = [\vartheta_T, y_T]$.
    \item Visual: Described more in Section \ref{camera}, the visual input is composed of 4 components: information if the cable is visible on the right side of the gripper, how confident is the angle, the angle, and the $y$ position of of the cable in relation to the gripper: $O_{V} = [v_V, q_V, \vartheta_V, y_V]$.
\end{itemize}

\subsubsection{Actions}
As the gripper is supposed to move freely in the $x$-$y$ space, our action array is composed of target displacements in these directions. Apart from that, the agent is able to modify the closing force of the grip hence the array of action is composed of: $A = [\Delta x_g, \Delta y_g, c_G]$. As the time step is constant in the simulation (set to 0.01 s) these $\Delta x$ and $\Delta y$ values directly correspond to velocities of the gripper. The simulator allows to set how many time steps are executed after each action step, in our case this was set to 8. We limit these values (by scaling the output of the agent) to keep the speed of the gripper in a feasible range for the real robot and also to assure better RL training. After some preliminary experiments we chose maximal values of $\Delta x$ and $\Delta y$ to be 0.0025, which corresponds to the velocity of 0.25 m/s.

\subsubsection{Reward} \label{reward}
Our reward can be represented by partial rewards and defined as:
\[
    R_t = 
\begin{cases}
    R_{move} + R_{end},& \text{if } n_h \neq 0\\
    P_{fall},          & \text{if }  n_h = 0
\end{cases}
\]
where $R_{move}$ is a reward for moving towards the end of the DLO; $R_{end}$ is the reward for being close to the end of the rope; $P_{fall}$ is the penalty for dropping it; $n_h$ is a number of particles being held by the gripper.

The partial rewards are defined as follows:
$$
R_{move} = \alpha_{move}(d_t - d_{t-1})
$$
where $\alpha_{move}$ is the weight of that reward, $d_t$ is the distance (in meters) at the current time step $t$. In the simulation, it is calculated as:
$$
d_t = \frac{p_i L_c}{n_c}
$$
where $p_i$ is the average value of indexes of particles being held by the gripper, $L_c$ is the length of the rope, and $n_c$ is number of particles that the rope is made of.

$R_{end}$ is given only when the gripper is closer than 20 particles from the end of the object and it is increasing linearly when fingers approach the end\footnote{Quadratic function was also tested.}.
\[
    R_{end} = 
\begin{cases}
    \alpha_{end}(p_i+20-n_c),& \text{if } p_i > n_c-20\\
    0,          & \text{otherwise}
\end{cases}
\]

$P_{fall}$ is a constant value,, together with $\alpha_{move}$ and $\alpha_{end}$, was chosen through a hyperparameter search. These values were set to $P_{fall} = - 0.5$, $\alpha_{move} = 10$, and $\alpha_{end} = \frac{1}{20}$.

\subsubsection{Evaluation Metrics} \label{metrics}
We evaluate the performance of our model using several metrics. One of them is to classify each of the completed episodes in one of the categories:
\begin{itemize}
    \item \textit{Hold the end} -- the gripper follows the DLO till its end, it stays there and holds the object. This is the goal behaviour. We defined being at the end as the situation when the gripper holds any of the last 10 particles.
    \item \textit{Stop before} -- the gripper did not reach close to the end but it did not drop the object.
    \item \textit{Reach end but drop} -- the gripper reached the end of the DLO (last 10 particles) but failed to keep the object.
    \item \textit{Drop before} -- the gripper dropped the DLO earlier, without reaching its end.
\end{itemize}

Apart from this classification, two more metrics are used:
\begin{itemize}
    \item \textit{Time spent at the end}
    -- we check how long (i.e., how many time steps) the gripper spends at the end of the DLO (at any of 10 last particles).
    \item \textit{How far it goes}
    -- we check how close to the end of the DLO gripper reached (we measure this distance from the end of the object as the length of the rope is randomised). The distance is measured in particles.
\end{itemize}


\section{Experiment Setup}

We simplified our observations to scalars (e.g. angles and positions). This allows relatively simple assessment of what piece of information is useful. Ensuring that the observations are not influenced by the process of information extraction from the real images. 
This approach could potentially help us to avoid domain shift in the future Sim-to-Real transfer. It also makes this method more general, allowing its application with different types of sensors.

 \subsection{Simulation}
For a simulator we use Nvidia Flex -- a particle based simulation technique \cite{muller2007position,macklin2014unified}, wrapped in SoftGym \cite{pmlr-v155-lin21a} - which is a set of benchmarks. SofGym provides a set of simulated environments and agents, it uses PyFlex \cite{li2018learning} that provides Python interface for Nvidia Flex, and Gym \cite{Brockman2016}, a toolkit for developing and comparing reinforcement learning algorithms.

\begin{figure}
	\centering
	\includegraphics[width = 0.85\columnwidth]{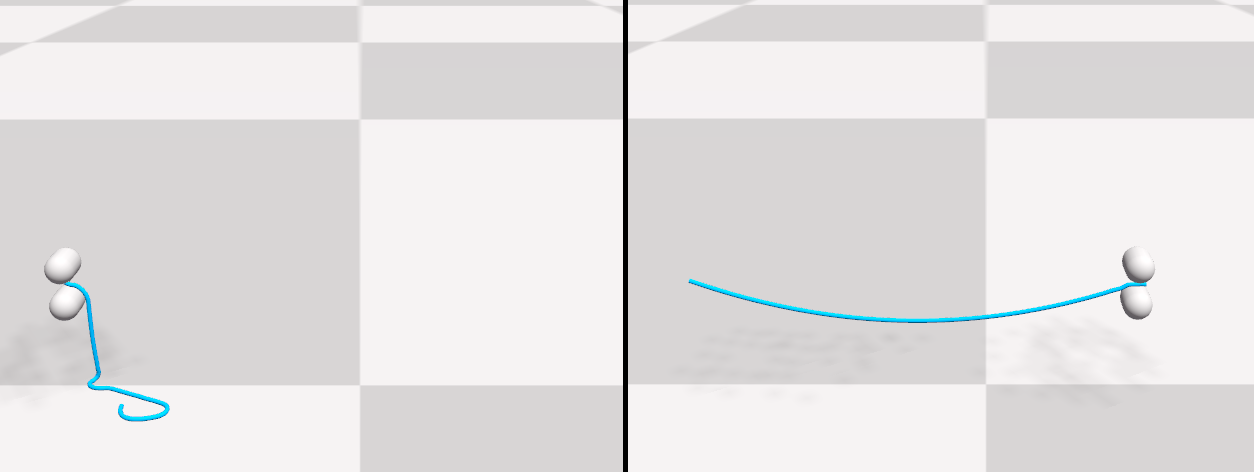}
	\caption{Frames from the simulated environment. On the left -- the beginning of the task where the cable falls freely; on the right -- the gripper finished the task and holds the cable's end.}
	\label{fig:Simulation}
\end{figure}

We constructed a new task and environment in SoftGym based on ``rope flatten'' task. 
We assumed the usage of a camera and a gripper with tactile sensors, based on that we amend the environment. We use two capsule-shape objects to simulate GelTip sensors \cite{gomes2020geltip}. We adjusted how the gripper can move and introduced a new way of gripping rope's (or a cable's) particles (Section \ref{gripper}).
The screenshots from the simulator, with an example of the rope configuration at the beginning and at the end of the task, can be seen in Fig.~\ref{fig:Simulation}.
\subsection{Tactile sensors} \label{tactile}
 
The sensor provides us with an angle $\vartheta_T$ of the cable with the gripper's $x$ axis as presented in Fig.~\ref{fig:State1}, and with the position of the cable in the $y$ direction (along the finger).
In the case of GelSight sensor, these angles and positions can be obtained using the algorithms described in \cite{yuan2017gelsight}.
We use particles' positions (between gripper's fingers) in the simulator to calculate these values (we fit the line to the centres of these particles and use its angle and its intersection with $y$ axis). In the case of the real sensor, the precision depends on the normal force, this is well illustrated in \cite{she2021cable} (for the GelSight sensor), where the controller is adapting the gripping force taking into account tactile quality. Bearing in mind that in some of the experiments presented herein we added a random noise -- proportional to the gripper closure -- to the measured angle and position.
$$\vartheta_T = \vartheta_{T,nom} + (1-c_G) \vartheta_{noise},$$
$$y_T = y_{T,nom} + (1-c_G) y_{noise},$$
where $\vartheta_{noise}$ and $y_{noise}$ are sampled from the normal distributions with different standard deviations (depending on assumed sensitivity); $\vartheta_{T,nom}$ and $y_{T,nom}$ are the nominal values. These are calculated based on the positions of the DLO particles between the grippers fingers.

There are two reasons why we imitate the finger-like sensors in our simulations. First, we are planning to use GelTip sensors in the future experiments with the robot. 
Second, due to the nature of the simulator -- a cable/rope is made of particles connected with springs in the Nvidia Flex environment -- rectangular-shape fingers are causing unwanted behaviour of the rope which is difficult to overcome (in general, the corner of the sensor was getting between particles and the gripper was pulling the cable), hence, more rounded shape is more appropriate for this simulation task.

\subsection{Camera}\label{camera}
\begin{figure}
	\centering
	\includegraphics[width = 0.63\columnwidth]{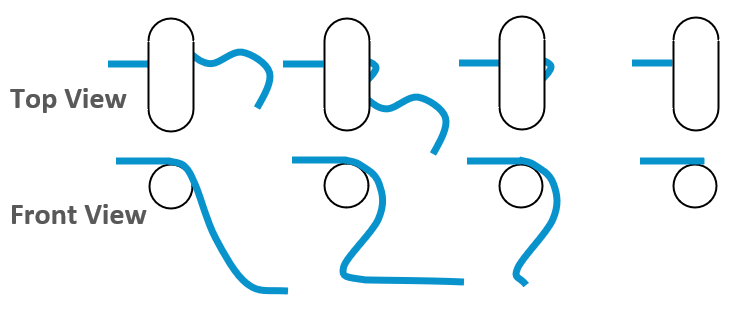}
	\caption{Possible occlusions in rope visibility caused by the gripper (capsule in the simulation).}
	\label{fig:State2}
\end{figure}

We assumed the usage of a top view camera. Similarly as with tactile data, we extract and use the position and angle of the rope from the camera image. We use the position and angle of the rope on the right from the gripper (in the direction of motion) as presented in Fig.~\ref{fig:State1} -- $\vartheta_v$ and $y_v$ variables. The camera's top view can be subjected to gripper and cable occlusions, and not always the position and angle of the cable are visible in the image as illustrated in Fig.~\ref{fig:State2} (top row). Hence, we also include the information if the cable is visible from the right -- $v_v$, and how far continuously it goes to the right -- the confidence about the angle -- $q_v$.



\subsection{Gripper} \label{gripper}
The gripper can move in $x$ and $y$ directions and close or open the grip, similarly as in \cite{she2021cable}. We implemented the process of gripping with variant gripping force in the simulated environment. This task is not trivial and in most simulations it is implemented by attaching the object being held to the gripper without taking into account possibility of sliding or such factors as friction. This was the case of the original environments implemented in SoftGym \cite{pmlr-v155-lin21a}. As the rope is in fact simulated as particles connected by springs, simple decreasing of the gap between two capsules (that we used to represent the gripper's fingers) was causing unnatural behaviours -- the gripper was getting caught between the particles holding them firmly.  

Instead of doing that, our intention was to modify the friction between the DLO and the fingers. This, on the other hand, was not straight forward because the friction parameters are global in the simulator. To this end, we modify the value of the inverse mass of the rope's particles that were between the fingers, and amended their positions according to the grippers' movement. This is similar to the approach in the original environments of the SoftGym, where the inverse mass was set to 0 and the position was set to follow the gripper --  causing that the particle was fully attached to it.

Assuming that the closing action is scaled between 0 and 1 and that the gripper changes the friction (or rather the inverse mass of hold particles) when that value is above 0.5 but the full closure appear at 0.9, the inverse mass can be expressed by the equation:
$$
w_p = max(2.25 w_{p,nom} - 2.5 w_{p,nom} max(c_G,0.5),0)
$$
where $w_{p,nom}$ is the nominal value of the inverses mass of DLO particles used in the simulator.

\subsection{Randomisation}
To allow our agent to operate in different environments we randomised variety of parameters in the simulation. This randomisation makes the agent to better generalize the task and could allow Real-to-Sim transfer even when the real environment, e.g., cable parameters, are quite different from these in the simulation.

We randomised:
\begin{itemize}
    \item Length of the cable (uniform, from 30 to 60 particles);
    \item DLO starting position -- we pick the rope in a random place and place it in a random location, we repeat that 4 times before each episode;
    \item DLO stretch stiffness (uniform, from 0.8 to 1.4);
    \item Bending stiffness (uniform, from 0.8 to 2.4);
    \item Friction coefficient (uniform, from 0.04 to 0.3);
\end{itemize}

The ranges of these parameters were chosen empirically in the simulator. Changing some of them in a bigger range would require to decrease the simulation step, which extends the training time. In these ranges the simulation was stable and at the same time we could observe different interactions between the gripper and the DLO.

\subsection{Training the agent}
In the training, we used a batch size of 128, learning rates for actor and critic of 0.001, 1,000 initial steps (with no agent updates), horizon length for each episode of 150 steps, and the maximum number of training steps was set to 50,000. 
The reward discount was set to $\gamma = 0.99$.

\section{Results}
To test the proposed methods and investigate how different sensory inputs contribute to the rope following task, we conducted multiple experiments. First, we compared the behavior of the agent and its performance when only one of the signals is provided. Next, we conducted similar experiments using two out of three inputs. Following that we performed the ablation studies where we trained the agent with all three inputs but we tested it excluding one of the signal. At the end, we check the agent's performance when different sensitivity of the tactile sensor is used -- different randomisation when the gripper is open.

All the results are collected in such a way that every 200 training steps we evaluate agent performance using 10 random environments (with a random DLO properties and configuration). We repeat the whole training 10 times for different random seeds (this way we train 10 different independent agents). The results presented in the paper are the average values from these 10 independent training sessions together with 95\% confidence margin. The curves were additionally smoothed using a windows of size 5 (the average of 5 following results).

\subsection{Training Performance When a Single Sensory Input is Provided}
\begin{figure}
	\centering
	\includegraphics[trim=0 60 0 40, clip,width = 1\columnwidth]{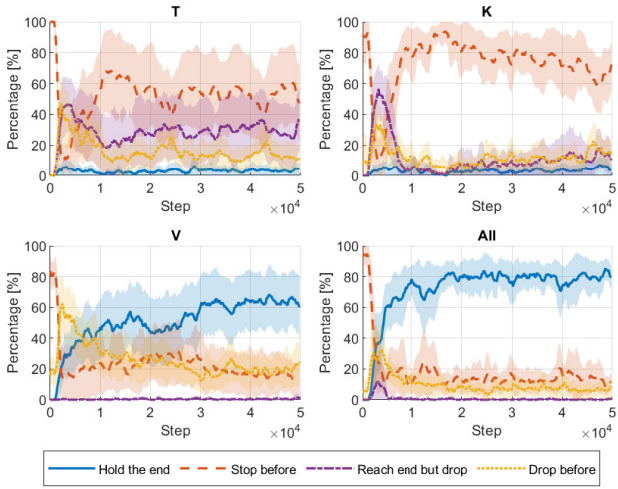}
	\caption{Episode outcome. Each subplot corresponds to a different sensory input: tactile, kinematic, vision, or all. The curves were collected from 10 independent training sessions, shading represents 95\% confidence range.}
	\label{fig:single}
\end{figure}

\begin{figure}
	\centering
	\includegraphics[trim=0 60 0 40, clip,width = 1\columnwidth]{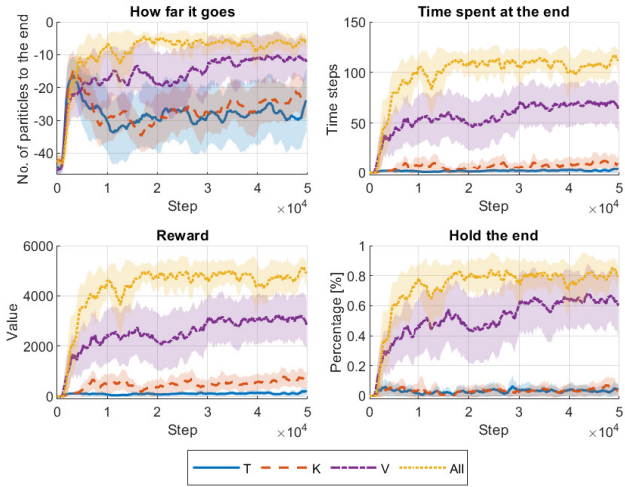}
	\caption{Agent's performance evaluation when one type of the input is used. Each subplot corresponds to a different metrics (episode reward and percentage of targeted outcome: \textit{Hold the end} were also included). The curves were collected from 10 independent training sessions, shading represents 95\% confidence range.}
	\label{fig:single_metrics2}
\end{figure}
We first analyze the results obtained when only one of the inputs was used (tactile, visual or kinematic). These are compared with the results when all three signals are provided together. Fig.~\ref{fig:single} shows the outcome of RL agent behaviour; how often the task was finalized in a most desired way: \textit{Hold the end}; or how frequently other 3 outcomes were observed (described more in Section.~\ref{metrics}). We can see that the behaviour of the agent changes significantly depending on the input signal. Only the vision input allows the model to stop and hold the DLO at the appropriate moment (\textit{Hold the end}). This was expected, as both \textit{T} and \textit{K} signals do not hold information that allows to identify the end of the object. This is, however, improved when all the signals are used together, in that case, the agent is able to obtain much better performance faster. As the kinematic signal does not provide any information about the DLO itself the agent prefers to not follow along the object and stops prematurely. Simulations with a tactile input show similar behaviour but the agent relatively often tends to follow the object till the end and drops it after that.

The results from particular modalities are compared more directly in the Fig.~\ref{fig:single_metrics2}. Each curve represents different type of inputs and each subplot shows different type of metrics. In the figure, we also included episode collective reward and the most desired outcome: \textit{Hold the end}. The best mean results for each modality are 12\%, 11\%, 77\% and 92\% (\textit{Hold the end}), respectively for \textit{T}, \textit{K}, \textit{V} and \textit{All}\footnote{As we mentioned before, the results in figures are smoothed for better readability so listed results might not be directly visible.}.

These results allow us to make a clear conclusion that the agent with a visual input outperforms the agents trained with other signal types. However, it is also clear that when other inputs are included this performance is improved. We can also see that both kinematic and tactile inputs help in sliding along DLO (\textit{How far it goes} subplot)\footnote{This edge following is fully possible with a tactile signal but it is not preferred by the agent due to the penalty for cable dropping which happens when it reaches the end.}.

\subsection{Training Performance When Two Sensory Inputs Are Provided}

\begin{figure}
	\centering
	\includegraphics[trim=0 60 0 40, clip,width = 1\columnwidth]{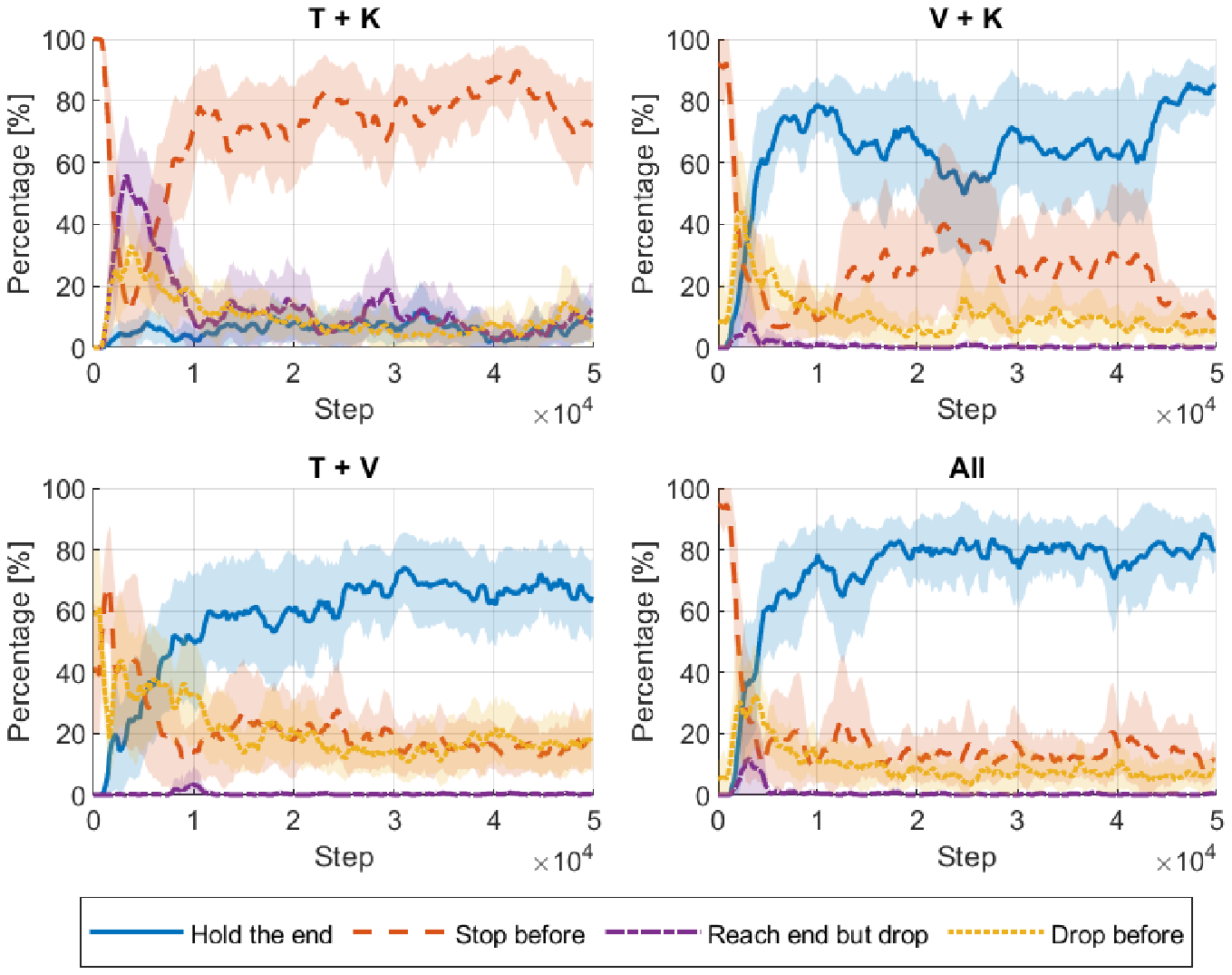}
	\caption{Episode outcome. Each subplot corresponds to a different combination of sensory input: tactile and kinematic, vision and kinematic, tactile and vision, or all. The curves were collected from 10 independent training sessions, shading represents 95\% confidence range.}
	\label{fig:dual}
\end{figure}

\begin{figure}
	\centering
	\includegraphics[trim=0 60 0 40, clip,width = 1\columnwidth]{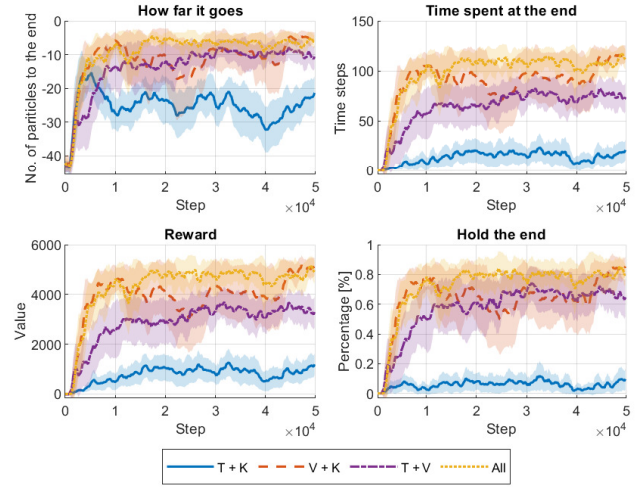}
	\caption{Agent's performance evaluation when two types of inputs are used. Each subplot corresponds to a different metrics (episode reward and percentage of targeted outcome: \textit{Hold the end} were also included). The curves were collected from 10 independent training sessions, shading represents 95\% confidence range.}
	\label{fig:dobule_metrics2}
\end{figure}

Figures \ref{fig:dual} and \ref{fig:dobule_metrics2} are created in the same manner as the figures in the earlier subsection. As expected, we can observe better performance of the training when two sensory inputs are used. When we compare the \textit{V} subplot from Fig.~\ref{fig:single} with subplots \textit{V + K} and \textit{T + V} presented here, we can see that each of the inputs (\textit{T} or \textit{K}) provides some improvement.

Again, we can see that the visual signal plays a crucial role and only when it is included in the input the agent is more often successfully performing \textit{Hold the end} behaviour.
The best mean results for each paired-modalities are 16\%, 89.0\%, 77\% and 92\% (\textit{Hold the end}), respectively for \textit{T + K}, \textit{V + K}, \textit{T + V} and \textit{All}.

In the case of \textit{How far it goes} metric, any combination that contains visual input (\textit{V + T} or \textit{T + V}) allows to achieve relatively high performance, similar to the one with \textit{All} signals.

\subsection{Ablation studies}

\begin{figure}
	\centering
	\includegraphics[trim=0 65 0 40, clip,width = 1\columnwidth]{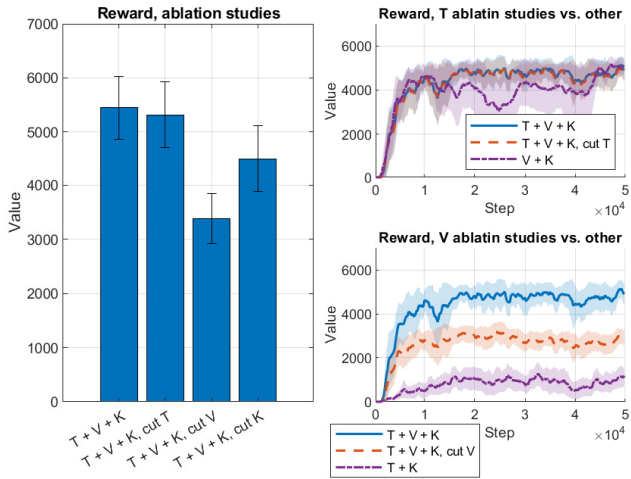}
	\caption{Performance of the agent when trained with all of the sensory inputs but some of the inputs were not provided in the testing phase. On the left subplot -- episode reward in a particular case; on the right -- comparing ablation studies (when tested without \textit{T} or \textit{V}) with all signal case and with the case when the model was trained from the beginning without one of the inputs. Error bars and shading correspond to 95\% confidence range.}
	\label{fig:ablation}
\end{figure}

We trained the agent using all of the inputs (\textit{T + V + K}), however, whenever the model was tested we did not provide one of the signals. The performance of the agent is presented in the left subplot in Fig.~\ref{fig:ablation}, where we show the mean value of collective episode reward. We can see that removal of \textit{V} signal caused the biggest drop in performance. Ablation of \textit{T} input, on the other hand, was insignificant and the results are almost the same as with that signal. This implies that the tactile input is in some way complementary.

To investigate the influence of the tactile signal, we present our ablation studies on a training development plot (Fig.~\ref{fig:ablation}, top right). The curve with removed signal is obtained in a way that the the agent is trained with all inputs but every 200 steps is tested with a tactile-free signal\footnote{We use the same agent for tactile-free and all signals tests, hence, the curve has a very similar characteristic. The difference in performance is more visible when checking other metrics.}. We can see that removal of \textit{T} input has practically no impact on the reward.
These results were compared with the session where tactile input was not used at all during the training. The results are interesting as we can see that the model which has access to additional tactile information in the training phase can learn faster and obtain better performance than the model trained without tactile signal \textit{T} (both agents are tested without \textit{T} input).
Possible explanation is that the DLO angle information (which is more certain in the case of tactile input) is useful in the training process to interpret and take advantage of the position information.
Similar observation is even more evident when the visual signal is removed (Fig.~\ref{fig:ablation}, bottom right). The agent that was trained with all signals performs much better without visual input compared to the model that was trained without vision.

\subsection{Tactile Sensitivity Study}

\begin{figure}
	\centering
	\includegraphics[trim=0 00 0 00, clip,width = 1\columnwidth]{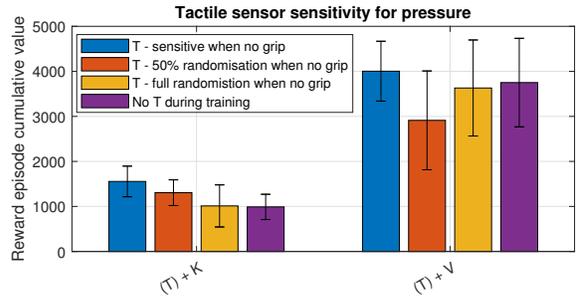}
	\caption{Performance of the agent (episode reward) when different sensitivity of the tactile sensor was assumed (sensitivity vary depending on the grasping force). Error bars correspond to 95\% confidence range.}
	\label{fig:sens}
\end{figure}

As described in Section \ref{tactile}, we took into account in our studies the impact of tactile sensor sensitivity. We included the random noise in the tactile input that was equal to zero when the gripper was using maximum gripping force and was increasing with gripper opening. Fig.~\ref{fig:sens} shows the results for different sensitivity and when the tactile input is not included at all. \textit{Full randomisation when no grip} corresponds to the situation where the standard deviation of aforementioned randomisation was set such high that angle should be practically not useful to estimate the real orientation of the rope ($\sigma$ in that case was set to 0.5) and position should be very unreliable (we set $\sigma$ to 0.002).
In that case, we can see that the agent was not able to learn to use that input (during the 50,000 steps training) -- the reward was as good as in the case of lack of tactile input.
However, we can observe improvement of the agent performance when partial randomisation was used if the \textit{K} input was provided together with \textit{T}. This shows that such less sensitive input can be used by the agent but only with the information about the gripper's closure, which allows to evaluate the reliability of the tactile signal.


\section{Conclusions and Future Works}

In this paper, we investigated the use of both vision and tactile inputs in completing a task of following deformable linear object. We introduced a benchmark in simulation and studied how an RL agent's behaviour can be boosted using visual-tactile fusion, compared to using single sensing inputs. We also conducted ablation studies where the agent, trained with a larger amount of signals, was tested with fewer inputs. Our results show the importance of multimodality and each sensing modality plays a different role in completing the task. In our particular task, the vision played a crucial role in finishing the task and finding the end of the cable. Without vision the agent prefers to finish the movement prematurely. We also see the importance of kinematic input which allows the agent to know where it is. Tactile input in some aspect was redundant with visual input. However, as we showed, it was also important for the agent to go further along the cable. The importance of the tactile signal is more significant when vision is not available (in real-life situations this can be quite common due to obstacles).

The results presented in this paper provide useful insights for future designs of tactile sensors and for deformable objects manipulation. The presented approach can provide guidance in the process of simulating tasks where sliding or touching of flexible materials is required. One of future works will be to adapt the trained agent on a real platform. Thanks to the usage of the distilled information, such transfer of knowledge should be much less affected by a domain shift. The research also has potential to be extended to more complex tasks, where we manipulate other objects (e.g., a cloth -- edge following) or we train the agent to achieve a different goal (e.g., wrapping a cable around the pin). 

\bibliographystyle{IEEEtran}
\bibliography{ref}

\end{document}